\documentclass[11pt,a4paper]{article}
\usepackage[table]{xcolor}
\usepackage[hyperref]{acl2020}
\usepackage{times}
\usepackage{booktabs}
\usepackage{latexsym}
\usepackage{arydshln}
\usepackage{graphicx}
\usepackage{url}
\usepackage[T1]{fontenc}
\usepackage{subfigure}
\usepackage{tabularx}
\usepackage{multirow}
\usepackage{multicol}
\usepackage{floatrow}
\usepackage{microtype}
\usepackage{amsmath}
\usepackage{amssymb}
\newcommand{\ignore}[1]{}

\usepackage{microtype}

\aclfinalcopy 

\title{Verb Knowledge Injection for Multilingual Event Processing}

\author{Olga Majewska$^1$, Ivan Vuli\'{c}$^{1}$, ~Goran Glava\v{s}$^{2}$, ~Edoardo M. Ponti$^{1,3}$, ~Anna Korhonen$^1$ \smallskip \\
$^1$Language Technology Lab, TAL, University of Cambridge, UK \\
$^2$ Data and Web Science Group, University of Mannheim, Germany \\
$^3$ Mila -- Quebec AI Institute, Montreal, Canada \\
$^1$\texttt {\{om304,iv250,ep490,alk23\}@cam.ac.uk} \\
$^2$\texttt {goran@informatik.uni-mannheim.de}
}

\date{}

\begin{document}
\maketitle
\begin{abstract}
In parallel to their overwhelming success across NLP tasks, language ability of deep Transformer networks, pretrained via language modeling (LM) objectives has undergone extensive scrutiny. While 
probing revealed that these models encode a range of syntactic and semantic properties of a language, 
they are still prone to fall back on superficial cues and simple heuristics to solve downstream tasks, rather than leverage deeper linguistic knowledge. In this paper, we target one such area of their deficiency, \textit{verbal reasoning}. We investigate whether injecting explicit information on verbs' semantic-syntactic behaviour improves the performance of LM-pretrained Transformers in event extraction tasks -- downstream tasks for which accurate verb processing is paramount. 
Concretely, we impart the verb knowledge from curated lexical resources into dedicated adapter modules (dubbed \textit{verb adapters}), allowing it to complement, in downstream tasks, the language knowledge obtained during LM-pretraining. We first demonstrate that injecting verb knowledge leads to performance gains in English event extraction. We then explore the utility of verb adapters for event extraction in other languages: we investigate (1) zero-shot language transfer with multilingual Transformers as well as (2) transfer via (noisy automatic) translation of English verb-based lexical constraints.
Our results show that the benefits of verb knowledge injection indeed extend to other languages, even when verb adapters are trained on noisily translated constraints. 



\end{abstract}

\section{Introduction}
\label{sec:intro}
Large Transformer-based encoders, pretrained with self-supervised language modeling (LM) objectives, form the backbone of state-of-the-art models for most Natural Language Processing (NLP) tasks \cite{devlin2019bert,yang2019xlnet,liu2019roberta}. Recent probing experiments showed that they implicitly extract a non-negligible amount of linguistic knowledge from text corpora in an unsupervised fashion \cite[\textit{inter alia}]{Hewitt:2019naacl,Vulic:2020emnlp,Rogers:2020arxiv}. In downstream tasks, however, they often rely on spurious correlations and superficial cues \citep{niven-kao-2019-probing} rather than a deep understanding of language meaning \citep{bender-koller-2020-climbing}, which is detrimental to both generalisation and interpretability \citep{mccoy-etal-2019-right}. 

In this work, we focus on a specific facet of linguistic knowledge, namely reasoning about events. For instance, in the sentence ``\textit{Stately, plump Buck Mulligan came from the stairhead, bearing a bowl of lather}'', an event of \textsc{coming} occurs in the past, with \textsc{Buck Mulligan} as a participant, simultaneously to an event of \textsc{bearing} with an additional participant, a \textsc{bowl}. Identifying tokens in the text that mention events and classifying the temporal and causal relations among them \citep{ponti-korhonen-2017-event} is crucial to understand the structure of a story or dialogue \citep{carlson2002rst,miltsakaki-etal-2004-penn} and to ground a text in real-world facts \citep{doddington2004automatic}. 

Verbs (with their arguments) are prominently used for expressing events (with their participants). Thus, fine-grained knowledge about verbs, such as the syntactic patterns in which they partake and the semantic frames they evoke, may help pretrained encoders to achieve a deeper understanding of text and improve their performance in event-oriented downstream tasks. Fortunately, there already exist expert-curated computational resources that organise verbs into classes based on their syntactic-semantic properties
\citep{jackendoff1992semantic,levin1993english}. In particular, here we consider VerbNet \citep{schuler2005verbnet} and FrameNet \citep{bick2011framenet} as rich sources of verb knowledge. 

Expanding a line of research on injecting external linguistic knowledge into (LM-)pretrained models \cite{peters2019knowledge,levine2019sensebert,lauscher-etal-2020-specializing}, we integrate verb knowledge into contextualised representations for the first time. We devise a new method to distill verb knowledge into dedicated \textit{adapter} modules \citep{houlsby2019parameter,pfeiffer2020adapterhub}, which reduce the risk of (catastrophic) forgetting of distributional knowledge and allow for seamless integration with other types of knowledge.

We hypothesise that complementing pretrained encoders through verb knowledge in such modular fashion should benefit model performance in downstream tasks that involve event extraction and processing.
%
We first put this hypothesis to the test in English monolingual event identification and classification tasks from the TempEval \citep{uzzaman-etal-2013-semeval} and ACE \citep{doddington2004automatic} datasets. Foreshadowing, we report modest but consistent improvements in the former, and significant performance boosts in the latter, thus verifying that verb knowledge is indeed paramount for deeper understanding of events and their structure.

Moreover, we note that expert-curated resources are not available for most of the languages spoken worldwide. Therefore, we also investigate the effectiveness of transferring verb knowledge across languages, and in particular from English to Spanish, Arabic and Mandarin Chinese. Concretely, we compare (1) zero-shot model transfer based on massively multilingual encoders and English constraints with (2) automatic translation of English constraints into the target language. Not only do the results demonstrate that both techniques are successful, but they also shed some light on an important linguistic question: to what extent can verb classes (and predicate--argument structures) be considered cross-lingually universal, rather than varying across languages \citep{valpal}?

Overall, our main contributions consist in 1) mitigating the limitations of pretrained encoders regarding event understanding by supplying verb knowledge from external resources; 2) proposing a new method to do so in a modular way through adapter layers; 3) exploring techniques to transfer verb knowledge to resource-poor languages. The gains in performance observed across four diverse languages and several event processing tasks and datasets warrant the conclusion that complementing distributional knowledge with curated verb knowledge is both beneficial and cost-effective.

\section{Verb Knowledge for Event Processing}
\label{sec:adapters}
\begin{figure*}[t!]
    \centering
    \includegraphics[width=0.7\textwidth]{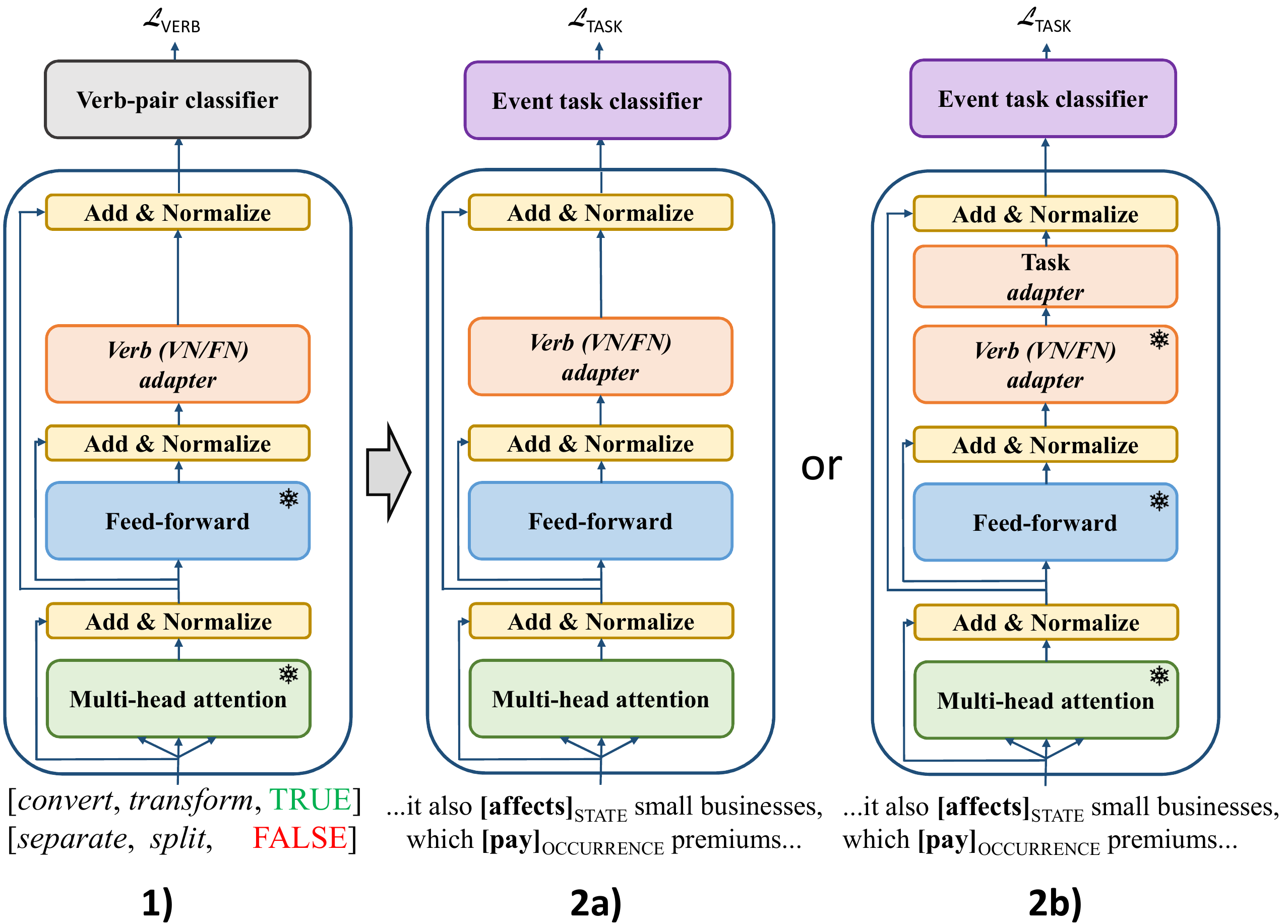}
    \vspace{-1.5mm}
    \caption{Framework for injecting verb knowledge into a pretrained Transformer encoder for event processing tasks. \textbf{1)} Dedicated \textit{verb adapter} parameters trained to recognise pairs of verbs from the same VerbNet (VN) class or FrameNet (FN) frame; \textbf{2)} Fine-tuning for an event extraction task (e.g., event trigger identification and classification \cite{uzzaman-etal-2013-semeval}): \textbf{a)} \textit{full fine-tuning} -- Transformer's original parameters and verb adapters both fine-tuned for the task; \textbf{b)} \textit{task adapter (TA) fine-tuning} -- additional \textit{task adapter} is mounted on top of \textit{verb adapter} and tuned for the task. For simplicity, we show only a single transformer layer; verb- and task-adapters are used in all Transformer layers. Snowflakes denote frozen parameters in the respective training step.}
    \label{fig:framework}
\end{figure*}

Figure~\ref{fig:framework} illustrates our framework for injecting verb knowledge from VerbNet or FrameNet and leveraging it in downstream event extraction tasks.
First, we inject the external verb knowledge, formulated as the so-called \textit{lexical constraints} \cite{Mrksic:2017tacl,ponti2019cross} (in our case -- verb pairs, see \S\ref{ss:lexica}), into a (small) additional set of \textit{adapter parameters} (\S\ref{ss:protocol}) \cite{houlsby2019parameter}. In the second step (\S\ref{ss:downstream}), we combine the language knowledge encoded by Transformer's original parameters and the verb knowledge from \textit{verb adapters} to solve a particular event extraction task. To this end, we either \textit{a)} fine-tune both sets of parameters (1. pretrained LM; 2. verb adapters) or \textit{b)} freeze both sets of parameters and insert an additional set of \textit{task-specific adapter parameters}. In both cases, the task-specific training is informed both by the general language knowledge captured in the pretrained LM, and the specialised verb knowledge, captured in the verb adapters.                 



\subsection{Sources of Verb Lexical Knowledge}
\label{ss:lexica}

Given the inter-connectedness between verbs' meaning and syntactic behaviour \cite{levin1993english,schuler2005verbnet}, we assume that refining latent representation spaces with semantic content and predicate-argument properties of verbs would have a positive effect on event extraction tasks that strongly revolve around verbs. Lexical classes, on the other hand, defined in terms of shared semantic-syntactic properties
provide a mapping between the verbs' senses and the morpho-syntactic realisation of their arguments \cite{jackendoff1992semantic,levin1993english}. The potential of verb classifications lies in their predictive power: for any given verb, a set of rich semantic-syntactic properties can be inferred based on its class membership. 
In this work, we explicitly harness this rich linguistic knowledge to help LM-pretrained Transformers in capturing regularities in the properties of verbs and their arguments, and consequently improve their ability to reason about events. 
We select two major English lexical databases -- VerbNet \cite{schuler2005verbnet} and FrameNet \cite{Baker:1998acl} -- as sources of verb knowledge at the semantic-syntactic interface, each representing a different lexical framework. Despite the different theories underpinning the two resources, their organisational units -- verb classes and semantic frames -- both capture regularities in verbs' semantic-syntactic properties.\footnote{Initially we also considered WordNet for creating verb constraints. While it provides records of verbs' senses and (some) semantic relations between them, WordNet lacks comprehensive information about the (semantic-)syntactic frames in which verbs participate. We thus believe that verb knowledge from WordNet would be less effective in downstream event extraction tasks than that from VerbNet and FrameNet.}

\vspace{1.4mm}
\noindent \textbf{VerbNet (VN)} \cite{schuler2005verbnet,Kipper:2006lrec} is the largest verb-focused lexicon currently available. It organises verbs into classes based on the overlap in their semantic properties and syntactic behaviour, building on the premise that a verb's predicate-argument structure informs its meaning \cite{levin1993english}. Each entry provides a set of thematic roles and selectional preferences for the verbs' arguments; it also lists the syntactic contexts characteristic for the class members. The classification is hierarchical, starting from broader classes and spanning several granularity levels where each subclass further refines the semantic-syntactic properties inherited from its parent class.\footnote{For example, within a top-level class `free-80', which includes verbs like \textit{liberate}, \textit{discharge}, and \textit{exonerate} which participate in a NP V NP PP.\textsc{theme} frame (e.g., \textit{It freed him of guilt.}), there exists a subset of verbs participating in a syntactic frame NP V NP S\_ING  (`free-80-1'), within which there exists an even more constrained subset of verbs appearing with prepositional phrases headed specifically by the preposition \textit{`from'} (e.g., \textit{The scientist purified the water from bacteria.}).} VerbNet's reliance on semantic-syntactic coherence as a class membership criterion means that semantically related verbs may end up in different classes because of differences in their 
combinatorial properties.\footnote{
E.g., verbs \textit{split} and \textit{separate} are members of two different classes with identical sets of arguments' thematic roles, but with discrepancies in their syntactic realisations (e.g., the syntactic frame NP V NP \textsc{apart} is only permissible for the `split-23.2' verbs: \textit{The book's front and back cover split apart}, but not  *\textit{The book's front and back cover separated apart}).} Although the sets of syntactic descriptions and corresponding semantic roles defining each VerbNet class are English-specific, the underlying notion of a semantically-syntactically defined verb class is thought to apply cross-lingually \cite{jackendoff1992semantic,levin1993english}, and its translatability has been demonstrated in previous work \cite{vulic2017cross,majewska2018investigating}. The current version of English VerbNet contains 329 main classes.

\vspace{1.4mm}
\noindent \textbf{FrameNet (FN)} \cite{Baker:1998acl}, in contrast to the syntactically-driven class divisions in VerbNet, is more semantically-oriented. Grounded in the theory of frame semantics \cite{fillmore1976frame,fillmore1977need,fillmore82:_frame}, it organises concepts according to semantic frames, i.e., schematic representations of situations and events, which they evoke, each characterised by a set of typical roles assumed by its participants. The word senses associated with each frame (FrameNet's lexical units) are similar in terms of their semantic content, as well as their typical argument structures.\footnote{For example, verbs such as \textit{beat}, \textit{hit}, \textit{smash}, and \textit{crush} evoke the `Cause\_harm' frame describing situations in which an \textit{Agent} or a \textit{Cause} causes injury to a \textit{Victim} (or their \textit{Body\_part}), e.g., \textit{A falling rock [Cause] CRUSHED the hiker's ankle [Body\_part]}, or \textit{The bodyguard [Agent] was caught BEATING the intruder [Victim]}. Note that frame-evoking elements do not need to be verbs; the same frame can also be evoked by nouns, e.g., \textit{strike} or \textit{poisoning}.} Currently, English FN includes the total of 1,224 frames and its annotations illustrate the typical syntactic realisations of the elements of each frame. Frames themselves are, however, semantically defined: this means that they may be shared even across languages with different syntactic properties (e.g., descriptions of transactions will include the same frame elements \textit{Buyer, Seller, Goods, Money} in most languages). Indeed, English FN has inspired similar projects in other languages: e.g., Spanish \cite{subirats2004spanish}, Swedish \cite{heppin2012rocky}, Japanese \cite{ohara2012semantic}, and Danish \cite{bick2011framenet}. 



\subsection{Training Verb Adapters}
\label{ss:adapterparams}


\label{ss:protocol}
\noindent \textbf{Training Task and Data Generation.}
 In order to encode information about verbs' membership in VN classes or FN frames into a pretrained Transformer, we devise an intermediary training task in which we train a dedicated VN-/FN-knowledge adapter (hereafter \textit{VN-Adapter} and \textit{FN-Adapter}). We frame the task as binary word-pair classification: we predict if two verbs belong to the same VN class or FN frame. We extract training instances from FN and VN independently. This allows for a separate analysis of the impact of verb knowledge from each resource.
 
We generate positive training instances by extracting all unique verb pairings 
from the set of members of each main VN class/FN frame (e.g., \textit{walk--march}), resulting in 181,882 positive instances created from VN and 57,335 from FN. 
We then generate $k = 3$ negative examples for each positive example in a training batch by combining controlled and random sampling. 
In controlled sampling, we follow prior work on semantic specialisation \cite{wieting2015paraphrase,glavas-vulic-2018-explicit,ponti2019cross,lauscher-etal-2020-specializing}. 
For each 
positive example $p=(w_1,w_2)$ in the training batch $B$, we create two negatives $\hat{p}_1=(\hat{w}_1,w_2)$ and $\hat{p}_2=(w_1,\hat{w}_2)$; $\hat{w}_1$ is the verb from batch $B$ other than $w_1$ that is closest to $w_2$ in terms of their cosine similarity in an auxiliary static word embedding space $X_{aux} \in \mathbb{R}^d$; conversely, $\hat{w}_2$ is the verb from $B$ other than $w_2$ closest to $w_1$. We additionally create one negative instance $\hat{p}_3 = (\hat{w}_1$,$\hat{w}_2$) by randomly sampling $\hat{w}_1$ and $\hat{w}_2$ from batch $B$, not considering $w_1$ and $w_2$. We make sure that negative examples are not present in the global set of all positive verb pairs from the resource.


 Similar to \citet{lauscher-etal-2020-specializing}, we tokenise each (positive \textit{and} negative) training instance into WordPiece tokens, prepended with sequence start token \texttt{[CLS]}, and with \texttt{[SEP]} tokens in between the verbs and at the end of the input sequence. 
 We consider the representation of the \texttt{[CLS]} token, $\mathbf{x}_\mathit{CLS} \in \mathbb{R}^h$ (with $h$ as the hidden state size of the Transformer), output by the last Transformer layer to be the latent representation of the verb pair, and feed it to a simple binary classifier:\footnote{We also experimented with sentence-level tasks for injecting verb knowledge, with target verbs presented in sentential contexts drawn from example sentences from VN/FN: we fed (a) pairs of sentences in a binary classification setup (e.g., \textit{Jackie \underline{leads} Rose to the store. -- Jackie \underline{escorts} Rose.}); and (b) individual sentences in a multi-class classification setup (predicting the correct VN class/FN frame). Both these variants with sentence-level input, however, led to weaker downstream performance.} 
 
 \vspace{-1em}
 
 \begin{equation}
    \hat{ \textbf{y}} = \textrm{softmax}(\textbf{x}_\textsc{cls}\textbf{W}_{cl} + \textbf{b}_{cl})
 \end{equation}
 
\noindent with $\textbf{W}_{cl} \in \mathbb{R}^{h\times2}$ and $\textbf{b}_{cl} \in \mathbb{R}^2$ as classifier's trainable parameters. We train by minimising the standard cross-entropy loss ($\mathcal{L}_\mathit{VERB}$ in Figure\,\ref{fig:framework}). 

\vspace{1.4mm}
\noindent \textbf{Adapter Architecture.} Instead of directly fine-tuning all parameters of the pretrained Transformer, we opt for storing verb knowledge in a separate set of adapter parameters, keeping the verb knowledge separate from the general language knowledge acquired in pretraining. This (1) allows downstream training to flexibly combine the two sources of knowledge, and (2) bypasses the issues with catastrophic forgetting and interference \cite{Hashimoto2017ajointmanytask,Autume:2019neurips}.
 

We adopt the adapter architecture of \citet{pfeiffer2020adapterfusion,pfeiffer2020mad} which exhibits comparable performance to more commonly used \citet{houlsby2019parameter} architecture, while being computationally more efficient. In each Transformer layer $l$, we insert a single adapter module ($Adapter_l$) after the feed-forward sub-layer. The adapter module itself is a two-layer feed-forward neural network with a residual connection, consisting of a down-projection $\textbf{D} \in \mathbb{R}^{h \times m}$, a GeLU activation \cite{Hendrycks:2020gelu}, and an up-projection $\textbf{U} \in \mathbb{R}^{m \times h}$, where $h$ is the hidden size of the Transformer model and $m$ is the dimension of the adapter: 

\vspace{-1.3em}

\begin{equation}
    Adapter_l(\textbf{h}_l, \textbf{r}_l) = \textbf{U}_l(\textrm{GeLU}(\textbf{D}_l(\textbf{h}_l))) + \textbf{r}_l
\end{equation}

\vspace{-0.1em}

\noindent where $\textbf{r}_l$ is the residual connection, output of the Transformer's feed-forward layer, and $\textbf{h}_l$ is the Transformer hidden state, output of the subsequent layer normalisation.

\subsection{Downstream Fine-Tuning for Event Tasks}
\label{ss:downstream}

With verb knowledge from VN/FN injected into the parameters of VN-/FN-Adapters, we proceed to the downstream fine-tuning for a concrete event extraction task. Tasks that we experiment with (see \S\ref{sec:eval}) are (1) token-level event trigger identification and classification and (2) span extraction for event triggers and arguments (a sequence labeling task). For the former, we mount a classification head -- a simple single-layer feed-forward softmax regression classifier -- on top of the Transformer augmented with VN-/FN-Adapters. For the latter, we follow the architecture from prior work \cite{m2019contextualized,wang2019adversarial} and add a CRF layer \cite{lafferty2001conditional} on top of the sequence of Transformer's outputs (for subword tokens), in order to learn inter-dependencies between output tags and determine the optimal tag sequence.              


For both types of downstream tasks, we propose and evaluate two different fine-tuning regimes: (1) \textit{full downstream fine-tuning}, in which we update both the original Transformer's parameters and VN-/FN-Adapters (see \textit{2a} in Figure~\ref{fig:framework}); and (2) \textit{task-adapter} (\textbf{TA}) \textit{fine-tuning}, where we keep both Transformer's original parameters and VN-/FN-Adapters frozen, while stacking a new trainable \textit{task adapter} on top of the VN-/FN-Adapter in each Transformer layer (see \textit{2b} in Figure~\ref{fig:framework}).

\subsection{Cross-Lingual Transfer}
\label{ss:clsri}

Creation of curated resources like VN or FN takes years of expert linguistic labour. Consequently, such resources do not exist for a vast majority of languages. Given the inherent cross-lingual nature of verb classes and semantic frames (see \S\ref{ss:lexica}), we investigate the potential for verb knowledge transfer from English to target languages, 
without any manual target-language adjustments. 
Massively multilingual Transformers, such as multilingual BERT (mBERT) \cite{devlin2019bert} or XLM-R \cite{conneau-etal-2020-unsupervised} have become the \textit{de facto} standard mechanisms for zero-shot (\textsc{zs}) cross-lingual transfer. We also adopt mBERT in our first language transfer approach: we fine-tune mBERT first on the English verb knowledge and then on English task data and then simply make task predictions for the target language input.  

Our second transfer approach, dubbed \textsc{vtrans}, is inspired by the work on cross-lingual transfer of semantic specialisation for static word embedding spaces \cite{Glavas:2019tutorial,ponti2019cross,wang-etal-2020-shikeblcu}.
\begin{table}[t]
\centering
\footnotesize{
\begin{tabularx}{1.0\columnwidth}{XXX}
\toprule
 & VerbNet  & FrameNet  \\ \midrule
English (EN) & 181,882 & 57,335 \\
Spanish (ES) & \phantom{x}96,300 & 36,623 \\
Chinese (ZH) & \phantom{x}60,365 & 21,815 \\
Arabic (AR) & \phantom{x}70,278 & 24,551 \\ 
 \bottomrule
\end{tabularx}
}%
\vspace{-1mm}
\caption{Number of positive training verb pairs in English, and in each target language obtained via the \textsc{vtrans} method (\S\ref{ss:clsri}).}
\label{tab:clsri}
\vspace{-2mm}
\end{table}
Starting from a set of positive pairs $P$ from English VN/FN, \textsc{vtrans} involves three steps: (1) automatic translation of verbs in each pair into the target language, (2) filtering of the noisy target language pairs by means of a relation prediction model trained on the English examples, and (3) training the verb adapters injected into either mBERT or target language BERT with target language verb pairs. For (1), we translate the verbs by retrieving their nearest neighbour in the target language from the shared cross-lingual embedding space, aligned using the Relaxed Cross-domain Similarity Local Scaling (RCSLS) model of \citet{joulin2018loss}. Such translation procedure is liable to error due to an imperfect cross-lingual embedding space as well as polysemy and out-of-context word translation. 
We dwarf these issues in step (2), where we purify the set of noisily translated target language verb pairs by means of a neural lexico-semantic relation prediction model, the Specialization Tensor Model \cite{glavas-vulic-2018-discriminating}, here adjusted for binary classification. We train the STM for the same task as verb adapters during verb knowledge injection (\S\ref{ss:adapterparams}): to distinguish (positive) verb pairs from the same English VN class/FN frame from those from different VN classes/FN frames. In training, the input to STM are static word embeddings of English verbs taken from a shared cross-lingual word embedding space. We then make predictions in the target language by feeding vectors of target language verbs (from noisily translated verb pairs), taken from the 
same cross-lingual word embedding space, as input for STM (we provide more details on STM training in Appendix \ref{app:stmparams}).
%
Finally, in step (3), we retain only the target language verb pairs identified by STM as positive pairs and perform \textit{direct} monolingual FN-/VN-Adapter training in the target language, following the same protocol used for English, as described in \S\ref{ss:protocol}.

\section{Experimental Setup}
\label{sec:eval}

\noindent \textbf{Event Processing Tasks and Data.}
In light of the pivotal role of verbs in encoding the unfolding of actions and occurrences in time, as well as the nature of the relations between their participants, sensitivity to the cues they provide is especially important in event processing tasks. There, systems are tasked with detecting that \textit{something happened}, identifying \textit{what type} of occurrence took place, as well as what \textit{entities} were involved. Verbs typically act as the organisational core of each such event schema,\footnote{Event expressions are not, however, restricted to verbs: adjectives, nominalisations or prepositional phrases can also act as event triggers (consider, e.g., \textit{Two weeks after the \underline{murder} took place...}, \textit{Following the recent \underline{acquisition} of the company's assets...}).} carrying a lot of semantic and structural weight. Therefore, a model's grasp of verbs' properties should have a bearing on ultimate task performance. Based on this assumption, we select event extraction and classification as suitable evaluation tasks to profile the methods from \S\ref{sec:adapters}.

These tasks and the corresponding data are based on the two prominent frameworks for annotating event expressions: TimeML \cite{pustejovsky2003timeml,pustejovsky2005specification} and the Automatic Content Extraction ({ACE}) \cite{doddington2004automatic}. First, we rely on the TimeML-annotated corpus from \textit{TempEval} tasks \cite{verhagen2010semeval,uzzaman-etal-2013-semeval}, which targets automatic identification of temporal expressions, events, and temporal relations. Second, we use the ACE dataset which provides annotations for entities, the relations between them, and for events in which they participate in newspaper and newswire text. We provide more derails about the frameworks and their corresponding annotation schemes in the Appendix~\ref{app:frameworks}.

\vspace{1.3mm}
\noindent \textbf{Task 1: Trigger Identification and Classification (TempEval).}
We frame the first event extraction task as a token-level classification problem, predicting whether a token triggers an event and assigning it to one of the following event types: OCCURRENCE (e.g., \textit{died, attacks}), STATE (e.g., \textit{share, assigned}), Reporting (\textit{e.g., announced, said}), I-ACTION (e.g., \textit{agreed, trying}), I-STATE (e.g., \textit{understands, wants, consider}), ASPECTUAL (e.g., \textit{ending, began}), and PERCEPTION (e.g., \textit{watched, spotted}).\footnote{E.g., in the sentence: \textit{``The rules can also \underline{affect} small businesses, which sometimes \underline{pay} premiums tied to employees' health status and claims history.''}, \textit{affect} and \textit{pay} are event triggers of type STATE and OCCURRENCE, respectively.}
We use the TempEval-3 data for English and Spanish \cite{uzzaman-etal-2013-semeval}, and the TempEval-2 data for Chinese \cite{verhagen2010semeval} (see Table \ref{tab:datasplits} for dataset sizes). 

\begin{table}
\footnotesize{

\begin{tabularx}{1.0\columnwidth}{lXXX}
\toprule
 &  & Train & Test \\ \midrule
\textbf{TempEval} & English & 830,005 & 7,174 \\
 & Spanish & \phantom{x}51,511 & 5,466 \\
 & Chinese & \phantom{x}23,180 & 5,313 \\
\textbf{ACE} & English & \phantom{xxx,}529 & \phantom{x,x}40 \\
 & Chinese & \phantom{xxx,}573 & \phantom{x,x}43 \\
 & Arabic & \phantom{xxx,}356 & \phantom{x,x}27 \\ 
 \bottomrule
\end{tabularx}}
\vspace{-1.5mm}
\caption{Number of tokens (TempEval) and documents (ACE) in the training and test sets.}
\label{tab:datasplits}
\vspace{-1.5mm}
\end{table}

\vspace{1.3mm}
\noindent \textbf{Task 2: Trigger and Argument Identification and Classification (ACE). }
In this sequence-labeling task, we detect and label event triggers and their arguments, with four individually scored subtasks: (i) trigger identification, where we identify the key word conveying the nature of the event, and (ii) trigger classification, where we classify the trigger word into one of the predefined categories; (iii) argument identification, where we predict whether an entity mention is an argument of the event identified in (i), and (iv) argument classification, where the correct role needs to be assigned to the identified event arguments. We use the ACE data available for English, Chinese, and Arabic.\footnote{The ACE annotations distinguish 34 trigger types (e.g., \textit{Business:Merge-Org}, \textit{Justice:Trial-Hearing}, \textit{Conflict:Attack}) and 35 argument roles. Following previous work \cite{hsi-etal-2016-leveraging}, we conflate eight time-related argument roles - e.g., `Time-At-End', `Time-Before', `Time-At-Beginning' - into a single `Time' role in order to alleviate training data sparsity.}

Event extraction as specified in these two frameworks 
is a challenging, highly context-sensitive problem, where different words (most often verbs) may trigger the same type of event, and conversely, the same word (verb) can evoke different types of event schemata depending on the context.
Adopting these tasks as our experimental setup thus tests whether leveraging fine-grained curated knowledge of verbs' semantic-syntactic behaviour can improve models' reasoning about event-triggering predicates and their arguments. 


\vspace{1.5mm}
\noindent \textbf{Model Configurations.}
For each task, we compare the performance of the underlying ``vanilla'' BERT-based model (see \S\ref{ss:downstream}) against its variant with an added VN-Adapter or FN-Adapter\footnote{We also experimented with inserting both verb adapters simultaneously; however, this resulted in weaker downstream performance than adding each separately, a likely product of the partly redundant, partly conflicting information encoded in these adapters (see \S\ref{ss:lexica} for comparison of VN and FN).} (see \S\ref{ss:adapterparams}) in two regimes: (a) full fine-tuning, and (b) task adapter (TA) fine-tuning (see Figure~\ref{fig:framework} again). To ensure that any performance gains are not merely due to increased parameter capacity offered by the adapter module, we evaluate an additional setup where we replace the knowledge adapter with a randomly initialised adapter module of the same size (\textit{+Random}). Additionally, we examine the impact of increasing the capacity of the trainable task adapter by replacing it with a \textit{`Double Task Adapter'} (\textsc{2TA}), i.e., a task adapter with double the number of trainable parameters compared to the base architecture described in \S\ref{ss:adapterparams}.

\vspace{1.5mm}
\noindent \textbf{Training Details: Verb Adapters.} We experimented with $k \in \{2,3,4\}$ negative examples and the following combinations of controlled ($c$) and randomly ($r$) sampled negatives (see \S\ref{ss:adapterparams}): $k=2$ $[c c]$, $k=3$ $[c c r]$, $k=4$ $[c c r r]$. In our preliminary experiments we found the $k=3$ $[ccr]$ configuration to yield best-performing adapter modules. 
The downstream evaluation and analysis presented in \S\ref{sec:results} is therefore based on this setup.


Our VN- and FN-Adapters are injected into the cased variant of the BERT Base model.  Following \citep{pfeiffer2020adapterfusion}, we train the adapters for 30 epochs using the Adam algorithm \cite{kingma2014adam}, a learning rate of $1e-4$ and the adapter reduction factor of 16 \cite{pfeiffer2020adapterfusion}, i.e., $d=48$. Our batch size is 64, comprising 16 positive examples and $3\times16=48$ negative examples (since $k=3$). We provide more details on hyperparameter search in Appendix~\ref{app:hparams1}.

\vspace{1.5mm}
\noindent \textbf{Downstream Task Fine-Tuning.}
In downstream fine-tuning on Task 1 (TempEval), we train for 10 epochs in batches of size 32, with a learning rate $1e-4$ and maximum input sequence length of $T = 128$ WordPiece tokens. In Task 2 (ACE), in light of a greater data sparsity,\footnote{Most event types in ACE ($\approx70\%$) have fewer than 100 labeled instances, and three have fewer than 10 \cite{liu2018event}.} we search for an optimal hyperparameter configuration for each language and evaluation setup from the following grid: learning rate $l \in \{1e-5, 1e-6\}$ and epochs $n \in \{3, 5, 10, 25, 50\}$ (with maximum input sequence length of $T = 128$).




\vspace{1.5mm}
\noindent \textbf{Transfer Experiments.} 
For zero-shot (\textsc{zs}) transfer experiments, we leverage mBERT, to which we add the VN- or FN-Adapter trained on the English VN/FN data. We train the model on English training data available for each task and evaluate it on the test set in the target language. For the \textsc{vtrans} approach (see \S\ref{ss:clsri}), we use language-specific BERT models readily available for our target languages, and leverage target-language adapters trained on translated and automatically refined verb pairs. The model, with or without the target-language VN-/FN-Adapter, is trained and evaluated on the training and test data available in the language. We carry out the procedure for three target languages (see Table~\ref{tab:clsri}). 
We use the same negative sampling parameter configuration proven strongest in our English experiments ($k=3$ $[ccr]$).



\section{Results and Discussion}
\label{sec:results}
\begin{table*}[!t]
{\footnotesize
\begin{tabularx}{1.0\linewidth}{llXXXX|XXXX}
\toprule

 &  & \textbf{FFT} & +Random & +FN & +VN & \textbf{TA} & +Random & +FN & +VN \\ \midrule
\textbf{TempEval} & \textsc{T-ident\&class} & 73.6 & 73.5 & 73.6 & 73.6 & 74.5 & 74.4 & \textbf{75.0} & \textbf{75.2} \\
\midrule
\textbf{ACE} & \textsc{T-ident} & 69.3 & 69.6 & \textbf{70.8} & \textbf{70.3} & 65.1 & 65.0 & \textbf{65.7} & \textbf{66.4} \\
 & \textsc{T-class} & 65.3 & 65.5 & \textbf{66.7} & \textbf{66.2} & 58.0 & 58.5 & \textbf{59.5} & \textbf{60.2} \\
 & \textsc{ARG-ident} & 33.8 & 33.5 & 34.2 & \textbf{34.6} & \phantom{x}2.1 & \phantom{x}1.9 & \phantom{x}2.3 & \phantom{x}2.5 \\
 & \textsc{ARG-class} & 31.6 & 31.6 & \textbf{32.2} & \textbf{32.8} & \phantom{x}0.6 & \phantom{x}0.6 & \phantom{x}0.8 & \phantom{x}0.8 \\ 

 \bottomrule
\end{tabularx}}
\caption{Results on English TempEval and ACE test sets for full fine-tuning (\textbf{FFT}) setup and the task adapter (\textbf{TA}) setup. Provided are average $F_1$ scores over 10 runs, with statistically significant (paired \textit{t}-test;  $p<0.05$) improvements over both baselines marked in bold.}
\label{tab:en_results}
\vspace{-1mm}
\end{table*}
\begin{table*}[t!]
{\footnotesize
\begin{tabularx}{1.0\linewidth}{llXXXX!{\vrule width .5pt}XXXX}
\toprule
 &  & \multicolumn{1}{l}{\bf FFT} & \multicolumn{1}{l}{+Random} & \multicolumn{1}{l}{+FN} & \multicolumn{1}{l}{+VN} & \multicolumn{1}{l}{\bf TA} & \multicolumn{1}{l}{+Random} & \multicolumn{1}{l}{+FN} & \multicolumn{1}{l}{+VN} \\
 \midrule
\textbf{Spanish} & \textsc{mBERT-zs} & 37.2 & 37.2 & 37.0 & 36.6 & 38.0 & 38.0 & \textbf{38.6} & 36.5 \\
 & ES-BERT & 77.7 & 77.1 & 77.6 & 77.4 & 70.0 & 70.0 & \textbf{70.7} & \textbf{70.6} \\
 & ES-\textsc{mBERT} & 73.5 & 73.6 & \textbf{74.4} & \textbf{74.1} & 65.3 & 65.4  & 65.8 & \textbf{66.2} \\
 \midrule
\textbf{Chinese} & \textsc{mBERT-zs} & 49.9 & 49.9 & \textbf{50.5} & 47.9 & 49.2 & 49.5 & \textbf{50.1} & 48.2 \\
 & ZH-BERT & 82.0 & 81.6 & 81.8 & 81.8 & 76.2 & 76.3 & 75.9 & \textbf{76.9}\\
 & ZH-\textsc{mBERT} & 80.2  & 80.1  & 79.9 & 80.0 & 71.8 & 71.8 & 72.1 & 71.9 \\
 \bottomrule
 
\end{tabularx}}
\caption{Results on Spanish and Chinese TempEval test sets for full fine-tuning (\textbf{FFT}) and the task adapter (\textbf{TA}) set-up, for zero-shot (\textbf{\textsc{zs}}) transfer with \textbf{mBERT} and monolingual target language evaluation with language-specific BERT (\textbf{ES-BERT / ZH-BERT}) or mBERT (\textbf{\textsc{ES-mBERT} / \textsc{ZH-mBERT}}), with FN/VN adapters trained on \textsc{vtrans}-translated verb pairs (see \S\ref{ss:clsri}). $F_1$ scores are averaged over 10 runs, with statistically significant (paired \textit{t}-test;  $p<0.05$) improvements over both baselines marked in bold.}
\label{tab:tempeval-multi}
\vspace{-1mm}
\end{table*}

\begin{table*}[t!]
{\footnotesize
\begin{tabularx}{1.0\linewidth}{lllXXXX!{\vrule width .5pt}XXXX}
\toprule
&  &  & \multicolumn{1}{l}{\bf FFT} & \multicolumn{1}{l}{+Random} & \multicolumn{1}{l}{+FN} & \multicolumn{1}{l}{+VN} & \multicolumn{1}{l}{\bf TA} & \multicolumn{1}{l}{+Random} & \multicolumn{1}{l}{+FN} & \multicolumn{1}{l}{+VN} \\
 \midrule
\textbf{Arabic} & \textsc{mBERT-zs} & \textsc{T-ident} & 15.8 & 13.5 & \textbf{17.2} & 16.3 & 29.4 & 30.3 & \textbf{32.9} & \textbf{32.4} \\
& & \textsc{T-class} & 14.2 & 12.2 & \textbf{16.1} & \textbf{15.6} & 25.6 & 26.3 & \textbf{27.8} & \textbf{28.4} \\
& & \textsc{ARG-ident} & \phantom{x}1.2 & \phantom{x}0.6 & \phantom{x}\textbf{2.1} & \phantom{x}\textbf{2.7} & \phantom{x}2.0 & \phantom{x}3.3 & \phantom{x}3.3 & \phantom{x}3.6 \\
& & \textsc{ARG-class} & \phantom{x}0.9 & \phantom{x}0.4 & \phantom{x}1.5 & \phantom{x}\textbf{1.9} & \phantom{x}1.2 & \phantom{x}1.6 & \phantom{x}1.6 & \phantom{x}1.3 \\
\cmidrule{3-11}
 & AR-BERT & \textsc{T-ident} & 68.8 & 68.9 & \textbf{70.2} & 68.6 & 24.0 & 21.3 & \textbf{24.6} & 23.5 \\
& & \textsc{T-class} & 63.6 & 62.8 & \textbf{64.4} & 62.8 & 22.0 & 19.5 & \textbf{23.1} & 22.3 \\
& & \textsc{ARG-ident} & 31.7 & 29.3 & \textbf{34.0} & \textbf{33.4} & \multicolumn{1}{c}{--} & \multicolumn{1}{c}{--\phantom{xxx}} & \multicolumn{1}{c}{--} & \multicolumn{1}{c}{--} \\
& & \textsc{ARG-class} & 28.4 & 26.7 & \textbf{30.3} & \textbf{29.7} & \multicolumn{1}{c}{--} & \multicolumn{1}{c}{--\phantom{xxx}} & \multicolumn{1}{c}{--} & \multicolumn{1}{c}{--} \\
\midrule
\textbf{Chinese} & \textsc{mBERT-zs} & \textsc{T-ident} & 36.9 & 36.7 & \textbf{42.1} & 36.8 & 47.8 & 49.4 & \textbf{55.0} & \textbf{55.4} \\
& & \textsc{T-class} & 27.9 & 25.2 & \textbf{30.9} & \textbf{29.8} & 38.6 & 40.1 & \textbf{43.5} & \textbf{44.9} \\
& & \textsc{ARG-ident} & \phantom{x}4.3 & \phantom{x}3.1 & \phantom{x}\textbf{5.5} & \phantom{x}\textbf{6.1} & \phantom{x}5.1 & \phantom{x}6.0 & \phantom{x}\textbf{7.6} & \phantom{x}\textbf{8.4} \\
& & \textsc{ARG-class} & \phantom{x}3.9 & \phantom{x}2.7 & \phantom{x}\textbf{4.9} & \phantom{x}\textbf{5.2} & \phantom{x}3.5 & \phantom{x}4.7 & \phantom{x}\textbf{5.7} & \phantom{x}\textbf{7.1} \\
\cmidrule{3-11}
 & ZH-BERT & \textsc{T-ident} & 75.5 & 74.9 & 74.5 & 74.9 & 69.8 & 69.3 & 70.0 & 70.2 \\
& & \textsc{T-class} & 67.9 & 68.2 & 68.0 & 68.6 & 58.4 & 57.5 & \textbf{59.9} & \textbf{60.0} \\
& & \textsc{ARG-ident} & 27.3 & 26.1 & \textbf{29.8} & \textbf{28.8} & \multicolumn{1}{c}{--} & \multicolumn{1}{c}{--\phantom{xxx}} & \multicolumn{1}{c}{--} & \multicolumn{1}{c}{--} \\
& & \textsc{ARG-class} & 25.8 & 25.2 & \textbf{28.2} & \textbf{27.2} & \multicolumn{1}{c}{--} & \multicolumn{1}{c}{--\phantom{xxx}} & \multicolumn{1}{c}{--} & \multicolumn{1}{c}{--} \\
\bottomrule

\end{tabularx}}
\caption{Results on Arabic and Chinese ACE test sets for full fine-tuning (\textbf{FFT}) setup and task adapter (\textbf{TA}) setup, for zero-shot (\textsc{zs}) transfer with \textbf{mBERT} and \textsc{vtrans} transfer approach with language-specific BERT (\textbf{AR-BERT / ZH-BERT}) and FN/VN adapters trained on noisily translated verb pairs (\S\ref{ss:clsri}). $F_1$ scores averaged over 5 runs; significant improvements (paired \textit{t}-test; $p<0.05$) over both baselines marked in bold.}
\label{tab:ace-multi}
\vspace{-1mm}
\end{table*}

\subsection{Main Results}

\noindent \textbf{English Event Processing.}
Table \ref{tab:en_results} shows the performance on English Task 1 (TempEval) and Task 2 (ACE). First, we note that the computationally more efficient setup with a dedicated task adapter (TA) yields higher absolute scores compared to full fine-tuning (FFT) on TempEval. When the underlying BERT is frozen along with the added FN-/VN-Adapter, the TA is enforced to encode additional task-specific knowledge into its parameters, beyond what is provided in the verb adapter, which results in two strongest results overall from the +FN/VN setups. In Task 2, the primacy of TA-based training is overturned in favour of full fine-tuning. Encouragingly, boosts provided by verb adapters are visible regardless of the chosen task fine-tuning regime, that is, regardless of whether the underlying BERT's parameters remain fixed or not. We notice consistent statistically significant\footnote{We test significance with the Student's \textit{t}-test with a significance value set at $\alpha=0.05$ for sets of model $F_1$ scores.} improvements in the +VN setup, although the performance of the TA-based setups clearly suffers in argument (\textsc{arg}) tasks due to decreased trainable parameter capacity. Lack of visible improvements from the Random Adapter supports the interpretation that performance gains indeed stem from the added useful `non-random' signal in the verb adapters. 

\vspace{1.4mm}
\noindent \textbf{Multilingual Event Processing.}
Table~\ref{tab:tempeval-multi} compares the performance of zero-shot (\textsc{zs}) transfer and monolingual target-language training (via the \textsc{vtrans} approach) on TempEval in Spanish and Chinese. For both we see that the addition of the FN-Adapter in the TA-based setup boosts zero-shot transfer. Benefits of this knowledge injection extend to the full fine-tuning setup in Chinese, achieving the top score overall. 

In monolingual evaluation, we observe consistent gains from the added transferred knowledge (i.e., the \textsc{vtrans} approach) in Spanish, while in Chinese performance boosts come from the transferred VerbNet-style class membership information (\textsc{+VN}). These results suggest that even the noisily translated verb pairs carry enough useful signal through to the target language.
To tease apart the contribution of the language-specific encoders and transferred verb knowledge to task performance, we carry out an additional monolingual evaluation substituting the monolingual target language BERT with the massively multilingual encoder, trained on (noisy) target language verb signal (\textsc{ES-mBERT/ZH-mBERT}). Notably, although the performance of the massively multilingual model is lower than the language-specific BERTs in absolute terms, the addition of the transferred verb knowledge 
helps reduce the gap between the two encoders, with tangible gains achieved over the baselines in Spanish (see discussion in \S\ref{ss:discuss}).\footnote{Given that analogous patterns were observed in relative scores of mBERT and language-specific BERTs in monolingual evaluation on ACE (Task 2), for brevity we show the \textsc{vtrans} results with mBERT on TempEval only.} 

In ACE, the top performance scores are achieved in the monolingual full fine-tuning setting; as seen in English, keeping the full capacity of BERT parameters unfrozen noticeably helps performance.\footnote{This is especially the case in \textsc{arg} tasks, where the TA-based setup fails to achieve meaningful improvements over zero, even with extended training up to 100 epochs. Due to the computational burden of such long training, the results in this setup are limited to trigger tasks (after 50 epochs).} In Arabic, FN knowledge provides performance boosts across the four tasks and with both the zero-shot (\textsc{zs}) and monolingual (\textsc{vtrans}) transfer approaches, whereas the addition of the VN adapter boosts scores in \textsc{arg} tasks. The usefulness of FN knowledge extends to zero-shot transfer in Chinese, and both adapters benefit the \textsc{arg} tasks in the monolingual (\textsc{vtrans}) transfer setup.
Notably, in zero-shot transfer, we observe that the highest scores are achieved in the task adapter (TA) fine-tuning, where the inclusion of the knowledge adapters offers additional performance gains. Overall, however, the argument tasks elude the restricted capacity of the TA-based setup, with very low scores across the board.

\subsection{Further Discussion}
\label{ss:discuss}


\noindent \textbf{Zero-shot Transfer vs Monolingual Training.}
The results reveal a considerable gap between the performance of zero-shot transfer and monolingual fine-tuning. The event extraction tasks pose a significant challenge to the zero-shot transfer via mBERT, where downstream event extraction training data is in English; however, mBERT exhibits much more robust performance in the monolingual setup, when presented with training data for event extraction tasks in the target language -- here it trails language-specific BERT models by less than 5 points (see Table \ref{tab:tempeval-multi}). This is an encouraging result, given that LM-pretrained language-specific Transformers currently exist only for a narrow set of well-resourced languages: for all other languages -- should there be language-specific event extraction data -- one needs to resort to massively multilingual Transformers.   
%
%
What is more, mBERT's performance is further improved by the inclusion of transferred verb knowledge (the \textsc{vtrans} approach, see \S\ref{ss:clsri}): in Spanish, where the greater typological vicinity to English (compared to Chinese) renders direct transfer of semantic-syntactic information more viable, the addition of verb adapters (trained on noisy Spanish constraints) yields significant improvements both in the FFT and the TA setup. These results confirm the effectiveness of lexical knowledge transfer (i.e., the \textsc{vtrans} approach) observed in previous work \cite{ponti2019cross,wang-etal-2020-shikeblcu} in the context of semantic specialisation of static word embedding spaces.
%


\vspace{1.4mm}
\noindent \textbf{Double Task Adapter.}
The addition of a verb adapter increases the parameter capacity of the underlying pretrained model. To verify whether increasing the number of trainable parameters in TA cancels out the benefits from the frozen verb adapter, we run additional evaluation in the TA-based setup, but with trainable task adapters double the size of the standard TA (\textbf{2TA}). Promisingly, we see in Tables~\ref{tab:2ta-tempeval} and \ref{tab:2ta-ace} that the relative performance gains from FN/VN adapters are preserved regardless of the added trainable parameter capacity. As expected, the increased task adapter size helps argument tasks in ACE, where verb adapters produce additional gains. Overall, this suggests that verb adapters indeed encode additional, non-redundant information beyond what is offered by the pretrained model alone, and boost the dedicated task adapter in solving the problem at hand.

\begin{table}[t!]
{\footnotesize
\begin{tabularx}{1.0\columnwidth}{llXXX}
\toprule
 &  & \multicolumn{1}{l}{\bf 2TA} & \multicolumn{1}{l}{+FN} & \multicolumn{1}{l}{+VN} \\ \midrule
\textbf{English} & EN-BERT & 74.5 & 74.8 & 74.8 \\
\midrule
\textbf{Spanish} & \textsc{mBERT-zs} & 37.7 & \textbf{38.3} & 37.1 \\
 & ES-BERT & 73.1 & \textbf{73.6} & \textbf{73.6} \\
 \midrule
\textbf{Chinese} & \textsc{mBERT-zs} & 49.1 & \textbf{50.1} & 48.8 \\
 & ZH-BERT & 78.1 & 78.1 & \textbf{78.6} \\ \bottomrule
\end{tabularx}}
\caption{Results on TempEval for the Double Task Adapter-based approaches (\textbf{2TA}). Significant improvements (paired \textit{t}-test;  $p<0.05$) in bold.}
\label{tab:2ta-tempeval}
\vspace{-2mm}
\end{table}

\begin{table}[t!]
{\footnotesize
\begin{tabularx}{1.0\columnwidth}{lllXXX}
\toprule
& & & \multicolumn{1}{l}{\bf 2TA} & \multicolumn{1}{l}{+FN} & \multicolumn{1}{l}{+VN} \\ \midrule
\textbf{EN} & EN-BERT & \textsc{T-ident} & 67.5 & \textbf{68.1} & \textbf{68.9} \\
 & & \textsc{T-class} & 61.6 & \textbf{62.6} & \textbf{62.7} \\
 & & \textsc{ARG-ident} & \phantom{x}6.2 & \phantom{x}\textbf{8.9} & \phantom{x}\textbf{7.1} \\
 & & \textsc{ARG-class} & \phantom{x}3.9 & \phantom{x}\textbf{6.7} & \phantom{x}\textbf{5.0} \\
\midrule
\textbf{AR} & \textsc{mBERT-zs} & \textsc{T-ident} & 31.2 & \textbf{32.6} & 31.7 \\
&  & \textsc{T-class} & 26.3 & \textbf{27.1} & \textbf{29.3} \\
 & & \textsc{ARG-ident} & \phantom{x}5.9 & \phantom{x}6.0 & \phantom{x}\textbf{6.9} \\
 & & \textsc{ARG-class} & \phantom{x}3.9 & \phantom{x}4.1 & \phantom{x}4.3 \\
 \cmidrule{4-6}
& AR-BERT & \textsc{T-ident} & 40.6 & \textbf{42.3} & \textbf{43.0} \\
 & & \textsc{T-class} & 36.9 & \textbf{38.1} & \textbf{39.5} \\
 & & \textsc{ARG-ident} & \multicolumn{1}{c}{--} & \multicolumn{1}{c}{--} & \multicolumn{1}{c}{--} \\
 & & \textsc{ARG-class} & \multicolumn{1}{c}{--} & \multicolumn{1}{c}{--} & \multicolumn{1}{c}{--} \\
 \midrule
\textbf{ZH} & \textsc{mBERT-zs} & \textsc{T-ident} & 54.6 & \textbf{56.3} & \textbf{58.1} \\
  & & \textsc{T-class} & 45.6 & \textbf{46.2} & \textbf{46.9} \\
 & & \textsc{ARG-ident} & \phantom{x}9.2 & \textbf{10.8} & \textbf{11.3} \\
 & & \textsc{ARG-class} & \phantom{x}8.0 & \phantom{x}8.5 & \phantom{x}\textbf{9.9} \\
 \cmidrule{4-6}
 & ZH-BERT & \textsc{T-ident} & 72.3 & \textbf{73.1} & 72.0 \\
 & & \textsc{T-class} & 59.6 & \textbf{63.0} & \textbf{61.3} \\
 & & \textsc{ARG-ident} & \phantom{x}2.6 & \phantom{x}2.8 & \phantom{x}3.3 \\
 & & \textsc{ARG-class} & \phantom{x}2.3 & \phantom{x}2.6 & \phantom{x}2.9 \\
 \bottomrule
\end{tabularx}}
\vspace{-1mm}
\caption{Results on ACE for the Double Task Adapter-based approaches (\textbf{2TA}). Significant improvements (paired \textit{t}-test;  $p<0.05$) in bold.}
\label{tab:2ta-ace}
\vspace{-1.5mm}
\end{table}

\begin{table}[t!]
\footnotesize{
\begin{tabularx}{1.0\columnwidth}{lXXX}
\toprule
 & \textbf{FFT}+FN$_{ES}$ & \textbf{TA}+FN$_{ES}$ & \textbf{2TA}+FN$_{ES}$ \\ \midrule
ES-BERT & 78.0\,(+0.4) & 70.9\,(+0.2) &	73.8\,(+0.2) \\
 \bottomrule
\end{tabularx}}
\vspace{-0.5mm}
\caption{Results ($F_1$ scores) on Spanish TempEval for different configurations of Spanish BERT with added Spanish FN-Adapter (\textbf{FN$_{ES}$}), trained on clean Spanish FN constraints. Numbers in brackets indicate relative performance w.r.t. the corresponding setup with FN-Adapter trained on (a larger set of) noisy Spanish constraints obtained through automatic translation of verb pairs from English FN (\textsc{vtrans} approach).}
\label{tab:es-tempeval}
\vspace{-1.5mm}
\end{table}

\vspace{1.4mm}
\noindent \textbf{Cleanliness of Verb Knowledge.} 
Gains from verb adapters suggest that there is potential to find supplementary information within structured lexical resources that can support distributional models in tackling tasks where nuanced knowledge of verb behaviour is important. The fact that we obtain best transfer performance through noisy translation of English verb knowledge suggests that these benefits transcend language boundaries.

There are, however, two main limitations to the translation-based (\textsc{vtrans}) approach we used to train our target-language verb adapters (especially in the context of VerbNet constraints): (1) noisy translation based on cross-lingual semantic similarity 
may already break the VerbNet class membership alignment (i.e., words close in meaning may belong to different VerbNet classes due to differences in syntactic behaviour); and (2) the language-specificity of verb classes due to which they cannot be directly ported to another language without adjustments
due to the delicate language-specific interplay of semantic and syntactic information. This is in contrast to the proven cross-lingual portability of synonymy and antonymy relations shown in previous work on semantic specialisation transfer \cite{Mrksic:2017tacl,ponti2019cross}, which rely on semantics alone. 
In case of VerbNet, despite the cross-lingual applicability of a semantically-syntactically defined verb class as a lexical organisational unit, the fine-grained class divisions and exact class membership may be too English-specific to allow direct automatic translation. On the contrary, semantically-driven FrameNet lends itself better to cross-lingual transfer, given that it focuses on function and roles played by event participants, rather than their surface realisations (see \S\ref{ss:lexica}). Indeed, although FN and VN adapters both offer performance gains in our evaluation, the somewhat more consistent improvements from the FN-Adapter may be symptomatic of the resource's greater cross-lingual portability.

To quickly verify if noisy translation and direct transfer from English curb the usefulness of injected verb knowledge, we additionally evaluate the injection of \textit{clean} verb knowledge obtained from a small lexical resource available in one of the target languages -- Spanish FrameNet \cite{subirats2004spanish}. 
Using the procedure described in \S\ref{ss:protocol}, we derive 2,886 positive verb pairs from Spanish FN and train a Spanish FN-Adapter (on top of the Spanish BERT) with this (much smaller but clean) set of Spanish FN constraints. 
The results in Table~\ref{tab:es-tempeval} show that, despite having 12 times fewer positive examples for training the verb adapter compared to the translation-based approach, the `native' Spanish verb adapter outperforms its \textsc{vtrans}-based counterpart (Table \ref{tab:tempeval-multi}), compensating the limited coverage with gold standard accuracy. 
Nonetheless, the challenge for using native resources in other languages lies in their very limited availability and expensive, time-consuming manual construction process. Our results reaffirm the usefulness of language-specific expert-curated resources and their ability to enrich state-of-the-art NLP models. This, in turn, suggests that work on optimising resource creation methodologies merits future research efforts on a par with modeling work.

\section{Related Work}
\label{sec:related}
\subsection{Event Extraction}

The cost and complexity of event annotation requires robust transfer solutions capable of making fine-grained predictions in the face of data scarcity. Traditional event extraction methods relied on hand-crafted, language-specific 
features \cite{ahn2006stages,gupta2009predicting,llorens2010tipsem,hong-etal-2011-using,li-etal-2013-joint,glavavs2015construction} (e.g., POS tags, entity knowledge, morphological and syntactic information), which limited their generalisation ability and effectively prevented language transfer.

More recent approaches commonly resorted to word embedding input and neural text encoders such as recurrent nets \cite{nguyen2016joint,duan-etal-2017-exploiting,sha2018jointevent} and convolutional nets \cite{chen-etal-2015-event,nguyen-grishman-2015-event}, as well as graph neural networks \cite{nguyen2018graph,yan2019event} and adversarial networks \cite{hong2018self,zhang2018event}. Like in most other NLP tasks, most recent empirical advancements in event trigger and argument extraction tasks have been achieved through fine-tuning of LM-pretrained Transformer networks \cite{yang2019exploring,wang2019adversarial,m2019contextualized,wadden2019entity,liu2020event}. 

Limited training data nonetheless remains an obstacle, especially when facing previously unseen event types. The alleviation of such data scarcity issues has been attempted through data augmentation methods -- automatic data annotation \cite{chen2017automatically,zheng-2018-corpus,araki-mitamura-2018-open} and bootstrapping for training data generation \cite{ferguson2018semi,wang2019adversarial}. The recent release of the large English event detection dataset MAVEN \cite{wang2020maven}, with annotations of event triggers only, partially remedies for training data scarcity. MAVEN also demonstrates that even the state-of-the-art Transformer-based models fail to yield satisfying event detection performance in the general domain. The fact that it is unlikely to expect datasets of similar size for other event extraction tasks (e.g., event argument extraction) and especially for other languages only emphasises the need for external event-related knowledge and transfer learning approaches, such as the ones introduced in this work.       

Beyond event trigger (and argument)-oriented frameworks such as ACE and its light-weight variant ERE \cite{aguilar-etal-2014-comparison,song2015light}, several other event-focused datasets exist which frame the problem either as a slot-filling task \cite{grishman1996message} or an open-domain problem consisting in extracting unconstrained event types and schemata from text \cite{allan2012topic,minard2016meantime,araki-mitamura-2018-open,liu2019open}. Small domain-specific datasets have also been constructed for event detection in biomedicine \cite{kim2008text,thompson2009construction,buyko2010genereg,nedellec2013overview}, as well as literary texts \cite{sims2019literary} and Twitter \cite{ritter2012open,guo2013linking}.  

\subsection{Semantic Specialisation}

Representation spaces induced through self-supervised objectives from large corpora, be it the word embedding spaces  \cite{mikolov2013distributed,bojanowski2017enriching} or those spanned by LM-pretrained Transformers \cite{devlin2019bert,liu2019roberta}, encode only distributional knowledge, i.e., knowledge obtainable from large corpora. 
A large body of work focused on \textit{semantic specialisation} (i.e., refinement) of such distributional spaces by means of injecting lexico-semantic knowledge from external resources such as WordNet \cite{fellbaum1998wordnet}, BabelNet \cite{navigli-ponzetto-2010-babelnet} or ConceptNet \cite{liu2004conceptnet} expressed in the form of lexical constraints \cite[\textit{inter alia}]{faruqui-etal-2015-retrofitting,Mrksic:2017tacl,glavavs2018explicit,kamath2019specializing,lauscher-etal-2020-specializing}. 

\textit{Joint specialisation} models \cite[\textit{inter alia}]{yu-dredze-2014-improving,Nguyen:2017emnlp,lauscher-etal-2020-specializing,levine2019sensebert} train the representation space from scratch on the large corpus, but augment the self-supervised training objective with an additional objective based on external lexical constraints. \newcite{lauscher-etal-2020-specializing} add to the Masked LM (MLM) and next sentence prediction (NSP) pretraining objectives of BERT \cite{devlin2019bert} an objective that predicts pairs of synonyms and first-order hyponymy-hypernymy pairs, aiming to improve word-level semantic similarity in BERT's representation space. In a similar vein, \newcite{levine2019sensebert} add the objective that predicts WordNet supersenses. While joint specialisation models allow the external knowledge to shape the representation space from the very beginning of the distributional training, this also means that any change in lexical constraints implies a new, computationally expensive pretraining from scratch.     

\textit{Retrofitting and post-specialisation} methods \cite[\textit{inter alia}]{faruqui-etal-2015-retrofitting,Mrksic:2017tacl,vulic2018post,ponti2018adversarial,glavavs2019generalized,lauscher-etal-2020-common,wang2020kadapters}, in contrast, start from a pretrained representation space (word embedding space or a pretrained encoder) and fine-tune it using external lexico-semantic knowledge. \newcite{wang2020kadapters} fine-tune the pre-trained RoBERTa \cite{liu2019roberta} with lexical constraints obtained automatically via dependency parsing, whereas \newcite{lauscher-etal-2020-common} use lexical constraints derived from ConceptNet to inject knowledge into BERT: both adopt adapter-based fine-tuning, storing the external knowledge in a separate set of parameters. In our work, we adopt a similar adapter-based specialisation approach. However, focusing on event-oriented downstream tasks, our lexical constraints reflect verb class memberships and originate from VerbNet and FrameNet.

\section{Conclusion}
\label{sec:concl}
We have investigated the potential of leveraging knowledge about semantic-syntactic behaviour of verbs to improve the capacity of large pretrained models to reason about events in diverse languages. We have proposed an auxiliary pretraining task to inject information about verb class membership and semantic frame-evoking properties into the parameters of dedicated adapter modules, which can be readily employed in other tasks where verb reasoning abilities are key. We demonstrated that state-of-the-art Transformer-based models still benefit from the gold standard linguistic knowledge stored in lexical resources, even those with limited coverage. Crucially, we showed that the benefits of the information available in resource-rich languages can be extended to other, resource-leaner languages through translation-based transfer of verb class/frame membership information.

In future work, we will incorporate our verb knowledge modules into alternative, more sophisticated approaches to cross-lingual transfer to explore the potential for further improvements in low-resource scenarios. Further, we will extend our approach to specialised domains where small-scale but high-quality lexica are available, to support distributional models in dealing with domain-sensitive verb-oriented problems.

\section*{Acknowledgments}
This work is supported by the ERC Consolidator Grant LEXICAL: Lexical Acquisition Across Languages (no 648909) awarded to Anna Korhonen. The work of Goran Glavaš is supported by the Baden-Württemberg Stiftung (Eliteprogramm, AGREE grant).

\bibliography{anthology,references}
\bibliographystyle{acl_natbib}

\appendix

\clearpage
\section{Frameworks for Annotating Event Expressions}
\label{app:frameworks}
Two prominent frameworks for annotating event expressions are TimeML \cite{pustejovsky2003timeml,pustejovsky2005specification} and the Automatic Content Extraction (ACE) \cite{doddington2004automatic}. TimeML was developed as a rich markup language for annotating event and temporal expressions, addressing the problems of identifying event predicates and anchoring them in time, determining their relative ordering and temporal persistence (i.e., how long the consequences of an event last), as well as tackling contextually underspecified temporal expressions (e.g., \textit{last month, two days ago}). Currently available English corpora annotated based on the TimeML scheme include the  TimeBank corpus \cite{pustejovsky2003timeml}, a human annotated collection of 183 newswire texts (including 7,935 annotated \textsc{events}, comprising both punctual \textit{occurrences} and \textit{states} which extend over time) and the AQUAINT corpus, with 80 newswire documents grouped by their covered stories, which allows tracing progress of events through time \cite{derczynski2017automatically}. Both corpora, supplemented with a large, automatically TimeML-annotated training corpus are used in the TempEval-3 task \cite{verhagen-pustejovsky-2008-temporal,uzzaman-etal-2013-semeval}, which targets automatic identification of temporal expressions, events, and temporal relations.

The ACE dataset provides annotations for entities, the relations between them, and for events in which they participate in newspaper and newswire text. For each event, it identifies its lexical instantiation, i.e., the \textit{trigger}, and its participants, i.e., the \textit{arguments}, and the roles they play in the event. For example, an event type ``Conflict:Attack'' (``It could swell to as much as \$500 billion if we go to war in Iraq.''), triggered by the noun `war', involves two arguments, the ``Attacker'' (``we'') and the ``Place" (``Iraq''), each of which is annotated with an entity label (``GPE:Nation''). 

\section{Adapter Training: Hyperparameter Search}
\label{app:hparams1}
We experimented with $n \in \{10,15,20,30\}$ training epochs, as well as an early stopping approach using validation loss on a small held-out validation set as the stopping criterion, with a patience argument $p \in \{2,5\}$; we found the adapters trained for the full 30 epochs to perform most consistently across tasks. 

The size of the training batch varies based on the value of $k$ negative examples generated from the starting batch $B$ of positive pairs: e.g., by generating $k= 3$ negative examples for each of $8$ positive examples in the starting batch we end up with a training batch of total size $8+3*8=32$. We experimented with starting batches of size $B \in \{8,16\}$ and found the configuration $k=3$, $B=16$ to yield the strongest results (reported in this paper). 

\section{STM Training Details}
\label{app:stmparams}
We train the STM using the sets of English positive examples from each lexical resource (Table \ref{tab:clsri}). Negative examples are generated using controlled sampling (see \S\ref{ss:protocol}), using a $k=2$ $[cc]$ configuration, ensuring that generated negatives do not constitute positive constraints in the global set. We use the pre-trained 300-dimensional static distributional word vectors computed on Wikipedia data using the \textsc{fastText} model \cite{bojanowski2017enriching}, cross-lingually aligned using the RCSLS model of \newcite{joulin2018loss}, to induce the shared cross-lingual embedding space for each source-target language pair. The STM is trained using the Adam optimizer \cite{kingma2014adam}, a learning rate $l=1e-4$, a batch size of 32 (positive and negative) training examples, for a maximum of 10 iterations. We set the values of other training hyperparameters as in \newcite{ponti2019cross}, i.e., the number of specialisation tensor slices $K=5$ and the size of the specialised vectors $h=300$.




\end{document}